\documentclass[letterpaper, 10 pt, journal, twoside]{IEEEtran}
\usepackage{xcolor}
\usepackage{graphicx}
\usepackage{hyperref}
\usepackage{siunitx}
\usepackage{amsmath}

\begin{document}

\title{\LARGE \bf
A Data-Centric Approach For Dual-Arm Robotic Garment Flattening
}

\author{Li Duan$^{1}$ and Gerardo Aragon-Camarasa$^{1}$
\thanks{$^{1}$Li Duan and Gerardo Aragon-Camarasa are with the School of Computing Science,
        University of Glasgow, Glasgow, G12 8RZ, United Kingdom
        {\tt\small l.duan.1@research.gla.ac.uk} and {\tt\small Gerardo.AragonCamarasa@glasgow.ac.uk}}
        \thanks{Digital Object Identifier (DOI): see top of this page.}
}

\markboth{IEEE Robotics and Automation Letters. Draft Version. Accepted XX, XX}
{Duan \MakeLowercase{\textit{et al.}}: A Data-Centric Approach For Dual-Arm Robotic Garment Flattening}

\maketitle

\begin{abstract}
Due to the high dimensionality of object states, a garment flattening pipeline requires recognising the configurations of garments for a robot to produce/select manipulation plans to flatten garments. In this paper, we propose a data-centric approach to identify \textit{known configurations} of garments based on a \textit{known configuration} network (KCNet) trained on depth images that capture the \textit{known configurations} of garments and prior knowledge of garment shapes. In this paper, we propose a data-centric approach to identify the \textit{known configurations} of garments based on a \textit{known configuration} network (KCNet) trained on the depth images that capture the \textit{known configurations} of garments and prior knowledge of garment shapes. The \textit{known configurations} of garments are the configurations of garments when a robot hangs garments in the middle of the air. We found that it is possible to achieve 92\% accuracy if we let the robot recognise the common hanging configurations (the \textit{ known configurations}) of garments. We also demonstrate an effective robot garment flattening pipeline with our proposed approach on a dual-arm Baxter robot. The robot achieved an average operating time of 221.6 seconds and successfully manipulated garments of five different shapes.
\end{abstract}

\begin{IEEEkeywords}
Computer Vision for Automation, Deep Learning for Visual Perception, Visual Learning, AI-Enabled Robotics
\end{IEEEkeywords}

\section{Introduction}
\IEEEPARstart{R}{obotic} deformable-object manipulation remains a significant challenge in robotic research. Deformable object manipulation covers many areas of daily life: robot-assisted garment wearing for disability \cite{7759647}, robotic laundry \cite{6631181}, automated manufacturing, and robotic cooking \cite{6653967}. Mastering deformable object manipulation for robots is key to a more accessible future for the disability community and advanced manufacturing.

Deformable objects have two characteristics: the dimensionality of their object states (configurations) is high, changes in their object states (configurations) are instant, and irregular deformation patterns when a robot manipulates deformable objects. When deformable objects such as garments are flattened, folded, or sponges are gripped, their deformation patterns are irregular, making it challenging to find manipulating points. Researchers traditionally consider two ways to handle the challenge: model-centric and data-centric approaches. Model-centric approaches \cite{lin2015picking,zaidi2017model,yu2017haptic} focus on finding and defining specific models for objects, and manipulation plans are derived and updated by monitoring changes in configurations of those models. Data-centric \cite{balaguer2011combining,pignat2017learning,tsurumine2019deep} approaches are divided into two categories. First, deformation patterns are learned from large-scale datasets, which are used for identifying grasping/manipulating points on garments. Those grasping/manipulating points are then used for flattening or operating other manipulations on garments. Second, robots are trained to learn skills for manipulating objects via reinforcement and imitation learning.

There are several problems with model-centric approaches. Firstly, defining or finding black-box models that represent deformable objects is challenging. Deformable objects tend to have an infinite number of states, which means single or multiple black-box models cannot represent all possible object states. Model-centric approaches are usually validated in objects with simple deformations such as sponges \cite{tawbe2017acquisition}, towels \cite{yan2020learning} or fruits\cite{lin2015picking}. Complex, deformable objects such as garments are rarely employed because black-box models for representing object states would require an almost infinite database representing all possible object states of those deformable objects.  Secondly, models cannot be generalised, which means changing objects means updating models. Different objects have different characteristics (e.g. materials and textures), which means that black-box models are specific and object-restricted. A new model is needed for a new object, while knowledge or parameters may be invalidated, causing robotic manipulation tasks constrained to specific objects.

Similarly, limitations exist for data-centric approaches. Robots are usually trained in simulated environments for reinforcement and imitation learning approaches due to limitations in real environments. Robots may be rewarded with actions that are not feasible in real environments due to the range of constraints of orientations and positions for robotic grippers \cite{twardon2018learning}. Robots can take 100-200 epochs to be trained, which would take a significant time if trained in real environments \cite{matas2018sim}. While reward policies trained in simulated environments can potentially be invalid in real environments due to the differences in external factors such as illumination conditions and the imperfections of simulated objects \cite{mcconachie2018estimating}.

\begin{figure*}[t]
    \centering
    \includegraphics[width=\textwidth]{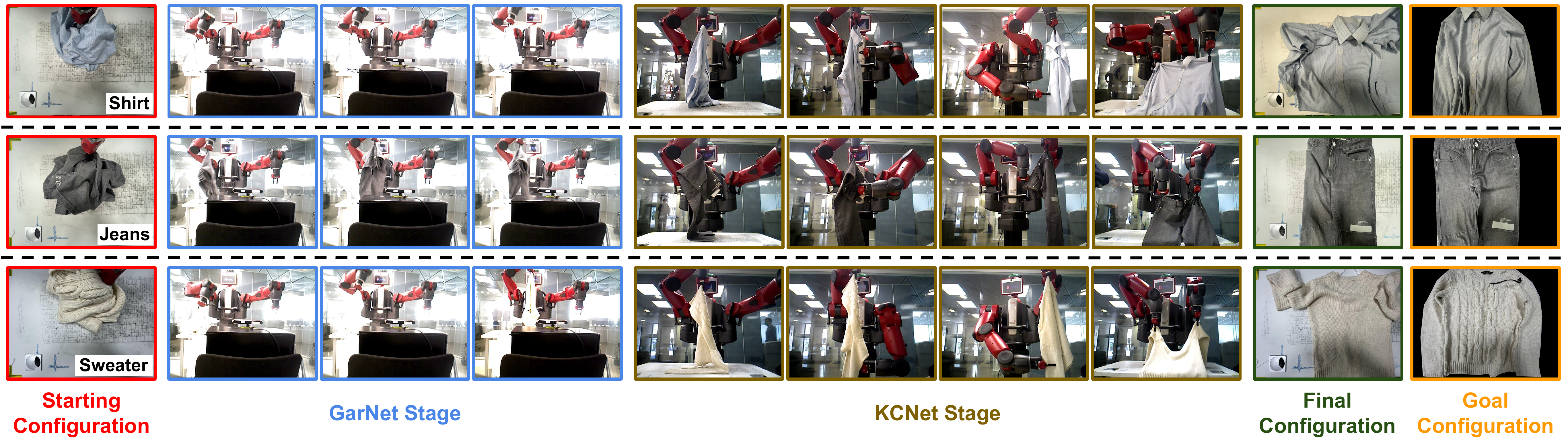}
    \caption{\textit{Robotic Demonstration Examples}: Garments with crumpled starting configurations are firstly recognised in their shapes by GarNet, and the Baxter robot recognises their \textit{known configurations} with KCNet from the depth images of their ‘known configurations’ and prior knowledge of their shapes. Then robot uses pre-designed manipulation plans to flatten those garments (the pre-designed grasping points are fine-tuned by the cloud-point method described in section \ref{subsec:grasping_points_locating_with_point_cloud}). From \textit{top} to \textit{bottom}: a shirt, a jean, a sweater}
    \label{fig:robotic_demonstration_examples}
\end{figure*}

In this paper, we propose a data-centric approach that aims to find \textit{known configurations} of garments and a robot that flattens garments with pre-designed manipulation plans. \textit{Known configurations} of garments are the configurations of hanging garments after a robot has grasped them from a given, random grasping point. A \textit{known configuration} network (KCNet) has been trained in this paper to recognise the \textit{known configurations} of garments with five different shapes. Compared with model-centric approaches, our approach does not require defining specific models for garments. We do not train a robot to learn manipulation skills with reinforcement or imitation learning. Instead, we propose that a robot select pre-designed manipulation plans based on the recognised \textit{known configurations} of garments. Traditional approaches such as \cite{yan2020learning} update manipulation plans by monitoring changes in the configurations of deformable objects in real-time using finite element methods, which is computationally costly. Using selected pre-designed manipulation plans, which will not be updated during manipulations, based on their recognised \textit{known configurations} is effective and fast. 

To validate our approach, we developed a robotic garment flattening pipeline, where a robot recognises the shapes of garments first and identifies the \textit{known configurations} of garments based on the recognised shapes. We hypothesise that \textit{A robot can effectively flatten garments of five garment shapes by first recognising the garments' shapes and identifying the \textit{known configurations} of garments based on the recognised shapes}. The contributions of this paper are threefold:

\begin{itemize}
\item We propose a robotic garment flattening pipeline, where a robot predicts the shapes of garments by continuously perceiving garments being grasped. Then, it recognises \textit{known configurations} of garments based on the predicted shapes and selects pre-designed garment manipulation plans to flatten these garments. Examples of successful manipulations can be found in Figure \ref{fig:robotic_demonstration_examples};
\item Our pipeline features an on-the-fly strategy. Traditional approaches \cite{qian2020cloth, li2015regrasping, triantafyllou2022garment} focus on accurately grasping points by learning features of different parts of garments in real-time. While in our approach, recognising the “known configurations” of garments makes it possible only to fine-tune the pre-designed grasping points to locate grasping points on the garments.
\item Our pipeline does not require sophisticated modelling for garments, as we learned the \textit{known configurations} of garments by training a network from a \textit{known configuration} dataset (i.e. a model-free and data-driven approach), resolving the challenges of modelling garments.
\end{itemize}


\section{Related Work \label{sec:related_work}}
\subsection{Model-centric approaches}
Miller \textit{et al.} \cite{miller2012geometric} proposed a simplified 2D polygon representation for garments, consisting of skeletal and folded models. Each model has four components: a landmark generator, a contour generator, a legal input (a set of parameters that ensure a model is reasonably defined) and a transformation generator. The simplified 2D representations are input to an algorithm which outputs the corresponding manipulation plans to fold garments. Their approach successfully enables a robot to manipulate garments of various shapes. However, fitting 2D polygon representations with garments need human intervention. Yan \textit{et al.} \cite{yan2020learning} introduced a model-based towel and rope flattening approach based on a contrastive estimation. A contrastive forward model (CFM) is proposed, which decides whether a predicted future object state belongs to the current object state with a specific action. In their experiments, images of towels and ropes with crumpled starting and goal configurations are input into the CFM, which outputs a sequence of object transformations between starting and goal configurations and corresponding actions. Their model achieved the best performance among other forward models. Still, they only validated their approach with simply-structured objects (towels and ropes), and the manipulations are time-consuming (about 15 minutes) for a towel flattening task. 

To find proper forces and manipulation plans for picking up deformable objects (fruits), Lin \textit{et al.} \cite{lin2015picking} proposed a “liftability test”, which is an algorithm to decide whether an object can be picked up with a given force by monitoring and analysing the deformations of the object. The objects were simulated by a finite element method (FEM) used for deformation monitoring and analysis. In their experiments, objects were simulated and modelled by FEM. FEM uses simulated meshes to model those objects and a simply-structured object (such as fruits) may take hundreds of meshes to be modelled, making FEM simulations computationally costly. If FEM simulations were to be adopted for complex structured objects (such as garments), more meshes would be needed to model the object and would require significant computational resources and delay real-time monitoring of its deformations. 

Yu \textit{et al.} \cite{yu2017haptic} introduced a model-based simulation approach to classifying the outcomes of robot-assisted dressing tasks. They trained hidden Markov models (HHMs) with simulated haptic data generated from an Nvidia PhysX simulator to classify the outcomes, tested the trained models in a real environment, and achieved high accuracy in classifying dressing tasks. One of the critical steps in their experiments is to optimise the Nvidia PhysX simulator with real-world haptic data such that the simulator can output simulated haptic data to train the HHMs. However, optimising the simulator during the training period takes a significant time (16.8 hours), decreasing the efficiency of their approach.

\subsection{Data-centric approaches: reinforcement and imitation learning}
McConachie \textit{et al.}\cite{mcconachie2018estimating} proposed a multiarmed bandit approach for deformable object manipulation. Instead of defining specific models for each type of deformable object, their approach features selecting models for manipulation that match deformable objects from the multi-armed bandit approach. They balanced explorations of models that fit objects and exploitations of high-utility models. Their approach addressed the limitations of models only for specific object types, and they validated their approach in simulated environments. Balaguer \textit{et al.} \cite{balaguer2011combining} introduced combining imitation and reinforcement learning to fold towels, featuring a “momentum fold” strategy where a robot swings towels to learn towel dynamics. They demonstrated that using the imitation technique in reinforcement learning can lead to faster algorithm convergence than reinforcement-only approaches. However, the initial configurations of the towels in their experiments are already flattened. Folding those towels is easier than folding towels from crumpled initial configurations, which are the main focus of this paper. 

\subsection{Data-centric approaches: finding grasping points}
Triantafyllou \text{et al.} \cite{triantafyllou2022garment} proposed to find grasping points of garments from their hanging configurations by finding common features in garments. They created a ‘dictionary’ of features of grasped garments: junctions, edges and folds. By matching features in the ‘dictionary’ with the features from grasped garments, they used pre-designed manipulation plans to unfold garments. They advanced state-of-art by not considering models for garments or robotic skills for manipulating garments with reinforcement-imitation learning. However, their approach would fail to manipulate unseen garments without a large-scale database to train a deep neural network to recognise hanging garments. Maitin-Shepard \textit{et al.} \cite{maitin2010cloth} proposed detecting the grasping points of towels by observing the grasped configurations of garments by dropping and re-grasping them until the grasped configurations are recognised. In their experiments, a robot has to find two grasping points of towels, meaning multiple dropping and re-grasping operations are needed. This process is time-consuming for towels, which are simpler structured than other garments such as jeans, shirts, sweaters and t-shirts. Their approach takes time to conduct the drop-and-re-grasp pipeline for complex garments until their grasped configurations are recognised.

Seita \textit{et al.}\cite{seita2019deep} presented an approach for detecting grasping points with a deep neural network for bed-making. The authors used a YOLO network trained on RGB images. They changed two layers in the YOLO network and trained the network with depth images capturing various bed-sheet configurations with ground-truth grasping points in a transfer learning way. The trained network showed high accuracy in finding grasping points for crumpled bed sheets. However, the network only supports bed sheets, and a larger database is needed to manipulate more deformable objects.

\begin{figure*}
    \centering
    \includegraphics[width=0.9\textwidth]{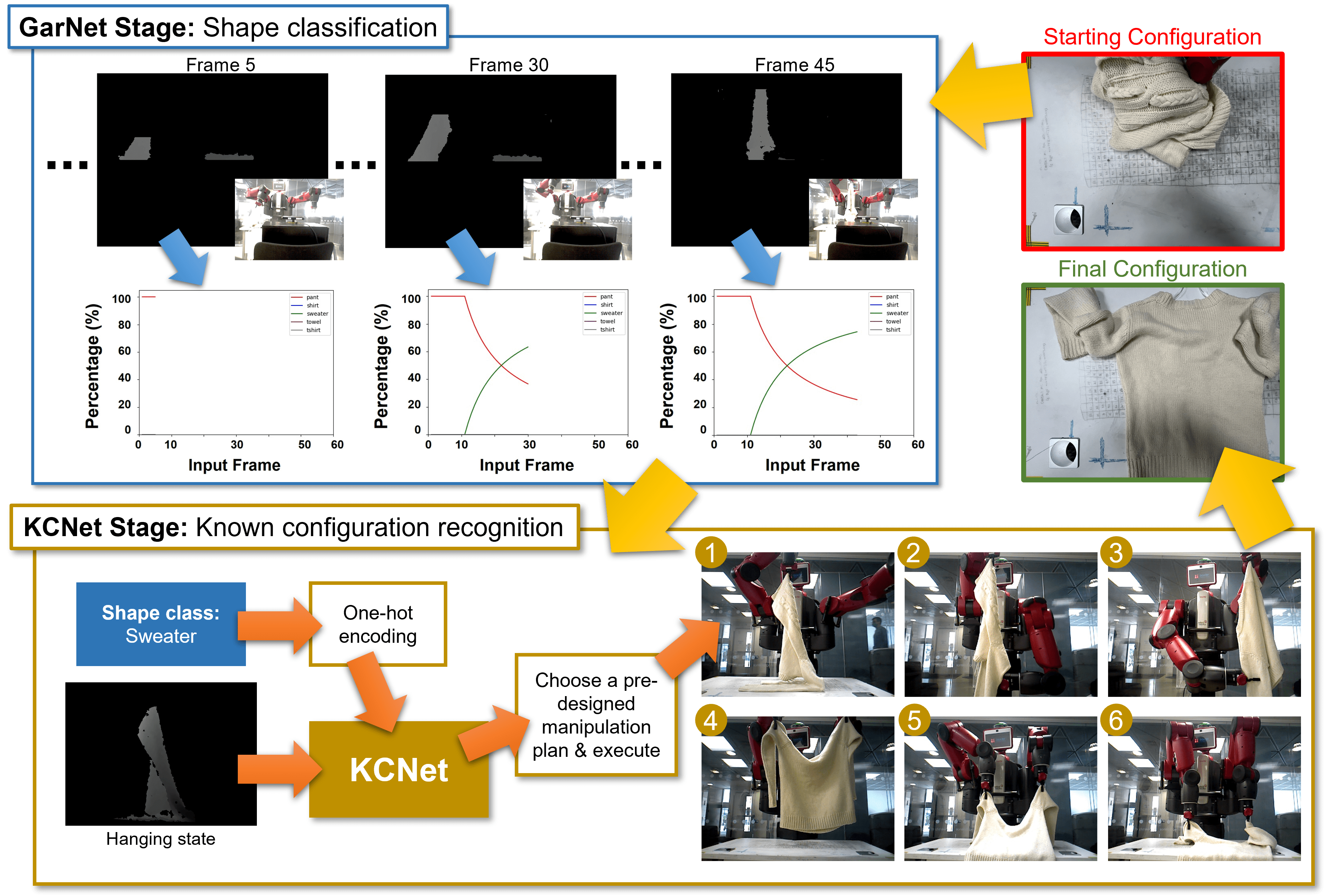}
    \caption{\textit{Full pipeline}: A sweater with a crumpled starting configuration is randomly grasped by the robot to a point above the table. The robot continuously perceives the sweater's motion trajectory to gain confidence in predicting the sweater’s shape. After, the robot recognises the \textit{known configuration} and then uses a pre-designed three-step manipulation plan to flatten the sweater. The pre-designed first and second grasping points are fine-tuned based on the closest point from the pre-defined grasping point and the hanging sweater.}
    \label{fig:main_figure}
\end{figure*}

Qian and Weng \textit{et al.} \cite{qian2020cloth} devised a grasping point estimation approach by proposing a network to segment garments into different regions on their depth images; including folds, wrinkles, edges and corners, which were segmented and coloured in different colours. Then a proposed grasping point location algorithm was used to find appropriate grasping points in the regions of edges and corners. Finally, the robot manipulated garments from the found grasping points. Their approach featured a garment region segmentation network, but only for towels. Edges and corners can be easily detected in towels, which can be difficult for other garments such as shirts.  A deep neural network is preferable for detecting other features to find grasping points.

Li \textit{et al.}\cite{li2015regrasping} proposed to recognise grasped configurations of garments for finding grasping points by matching 3D reconstructed mesh models of garments with simulated 3D mesh models in a database. A robot rotated garments to reconstruct their 3D models and re-grasps garments until the 3D models matched to 3D models in the database. The grasping points are found from matched 3D models in the database; then, a Baxter robot unfolds them. Their approach required rotations and re-grasps, which incurred in extra manipulation operations. In Li \textit{et al.} \cite{li2018model}, the authors advanced their previous approach in \cite{li2015regrasping} by proposing a feature extraction approach to match reconstructed 3D mesh models with the 3D mesh models in the database. However, rotations and re-grasps are still needed for matching.

In \cite{duan2022recognising}, we have proposed recognising grasped configurations of garments similar to \cite{triantafyllou2022garment,maitin2010cloth,li2015regrasping,li2018model} and using pre-design manipulation plans to flatten garments. Our previous approach features a deep neural network called a \textit{known configuration} network (KCNet) trained on grasped configurations (called \textit{known configurations}), which enables recognising \textit{known configurations} of unseen garments. Compared with previous work \cite{triantafyllou2022garment,maitin2010cloth,li2015regrasping,li2018model}, Our approach is more robust to unseen garments; however, due to training on garments of various shapes, the previous KCNet could not distinguish between shirts and sweaters due to their shape similarity. Our approach advances our previous approach by introducing prior knowledge on garments shapes, which helps identify the \textit{known configurations} of garments.

\section{Materials and Methodology \label{sec:methodology}}

\subsection{Flattening pipeline \label{subsec:full_pipeline}}

The flattening task in this paper is defined as the robot sliding the garment over the edge of a table after the garment has been unfolded in the air based on prior knowledge of its shape and \textit{known configuration}. Therefore, the proposed flattening pipeline consists of three core stages; shape prediction, known configuration prediction and manipulation stages as shown in Fig. \ref{fig:main_figure}.

The flattening starts with a robot grasping a garment from a table to a point above the table (approx. 1m from the table). During the motion trajectory, the robot predicts the garment's shape by continuously perceiving it as it is being translated to estimate its shape. For this, we employed a garment similarity network (ref. \ref{subsec:prior_knowledge_of_the_shapes_of_garments}). After the robot successfully predicts the garment shape, it predicts the garment's \textit{known configuration} (ref. Section \ref{subsec:robotic_garment_flattening_with_known_configuration_network}) using the depth image of the garment hanging state and the prior knowledge of the garment shape as inputs. A pre-designed manipulation plan matched with the recognised known configurations is used to flatten garments (ref. Section \ref{subsec:grasping_points_locating_with_point_cloud}).


\subsection{Garment's shape prediction \label{subsec:prior_knowledge_of_the_shapes_of_garments}}
The first part of our robotic garment flattening pipeline is about predicting the shapes of garments. The prior knowledge of the shapes of garments facilitates recognising the \textit{known configurations} of garments for matching manipulation plans to flatten garments. In our experiments, we employ our previous garment similarity network (GarNet) \cite{duan2021garnet} method to predict garment shapes. Compared with traditional approaches of using single images to predict garment shapes, GarNet enables a robot to continuously perceive the video frames of garments being grasped from a table to a point above the table. 

Specifically, the robot grasps a garment from a table to a point above the table (approx. $1m$ above the table). A sequence of depth images (as captured while the robot moves the garment from the table) are input into GarNet, which maps each image into a Garment Similarity Map (GSM) defined as a Garment Similarity Point (GSP). On the GSM, each garment shape learned by GarNet is clustered together, and the GSP from the input depth image is mapped into one of these shape clusters. After the robot perceives at least 20 images, if $80\%$ of GPSs from the input images belong to the same shape, the shape cluster class is used as the predicted shape for the grasped garment. Further details and experiments can be found in \cite{duan2021garnet}. We use a GarNet trained on depth images based on the experiment results from \cite{duan2021garnet}.

\subsection{Robotic garment flattening with known configuration network (KCNet) \label{subsec:robotic_garment_flattening_with_known_configuration_network}}
After garment shapes are successfully predicted by GarNet in section \ref{subsec:prior_knowledge_of_the_shapes_of_garments}, the robot will recognise the \textit{known configurations} of garments and match pre-designed manipulation plans with the recognised \textit{known configurations} to flatten garments. Figure \ref{fig:main_figure} shows the pipeline of the proposed robotic garment flattening pipeline.

We define a \textit{known configuration} as the configuration of a garment in a hanging state after being grasped from a different configuration. That is, the \textit{known configurations} of garments highly depend on grasping points because of gravity, regardless of their starting configurations, which means that the irregular and complex starting configurations of garments are converted into stable, constant \textit{known configurations}.

For this, we have adopted KCNet (described in \cite{duan2022recognising}) such that it can recognise the \textit{known configurations} of garments based not only on images of their \textit{known configurations} but also on prior knowledge of garment shapes, which GarNet predicted (Section \ref{subsec:prior_knowledge_of_the_shapes_of_garments}).

KCNet consists of a ResNet18 convolutional network and three linear networks. A KCNet consists of a ResNet18 convolutional network \cite{he2016deep} and three linear networks. An image of an \textit{Known Configuration} is an input to the KCNet to get its extracted features. The garment shape, as predicted by GarNet is encoded to a one-hot vector and then input into a linear network to get its latent representation. The feature extracted from the image and the latent representation from the one-hot vector of the predicted shape is concatenated, which is then input into a linear network to output the \textit{known configuration} of the garment. Similar to \cite{duan2022recognising}, our KCNet is trained with a Negative Log-Likelihood Loss (NLLLoss).

\subsection{Pre-designed manipulation plans \label{subsec:pre-designed_manipulation_plans}}

After the \textit{known configurations} of garments are successfully recognised from KCNet, the robot matches pre-designed manipulation plans with the recognised \textit{known configurations}. A pre-designed manipulation plan is a sequence of end-effector commands for the robot to flatten a garment. Each gripper command contains 16 parameters, including $XYZ$ coordinates of the right and left grippers, orientations (defined as a quaternion) of right and left grippers, a choice of grippers (left or right) and gripper status (open or close). On average, 18 gripper robotic actions are needed to flatten a garment. 

Similar to \cite{li2015regrasping} \cite{li2018model}, a manipulation plan follows a three-step manipulation rule: find the first grasping point, the second grasping point, and stretch and flatten garments. A \textit{known configuration} has a corresponding pre-designed manipulation plan. We have ten pre-designed manipulation plans for each shape of garments, totalling 50 manipulation plans as we have five shapes in our experiments.

\subsection{Fine-tuning Garment Grasping\label{subsec:grasping_points_locating_with_point_cloud}}
We pre-defined grasping points in our pre-designed manipulation plans, consisting of the first and second grasping points. However, garments deform irregularly, so those pre-designed grasping points must be fine-tuned for stable grasps during a flattening operation. Therefore, we proposed to locate grasping points by finding the closest point between the robot's gripper and the garment in a point cloud.

Firstly, our Xtion camera is placed in front of the robot for recognising \textit{known configurations} of garments. To bring the camera and the robot into alignment, we use Khan's \textit{et al.} \cite{7866616} hand-eye calibration approach. After the robot and camera coordinate systems are aligned, we capture point clouds of garments. From the captured point cloud, the points belonging to garments have a specific range of values because they have similar distances to the camera. Therefore, points belonging to garments can be easily identified and segmented from the points belonging to other objects (for example, the Baxter robot). Then, we find points that have the smallest distances with pre-designed grasping points and define them as fine-tuned grasping points used for grasping garments. We can ensure that grasping points are on garments and stable grasps can be generated on-the-fly in this way.

\begin{figure}
    \centering
    \includegraphics[width=0.5\textwidth]{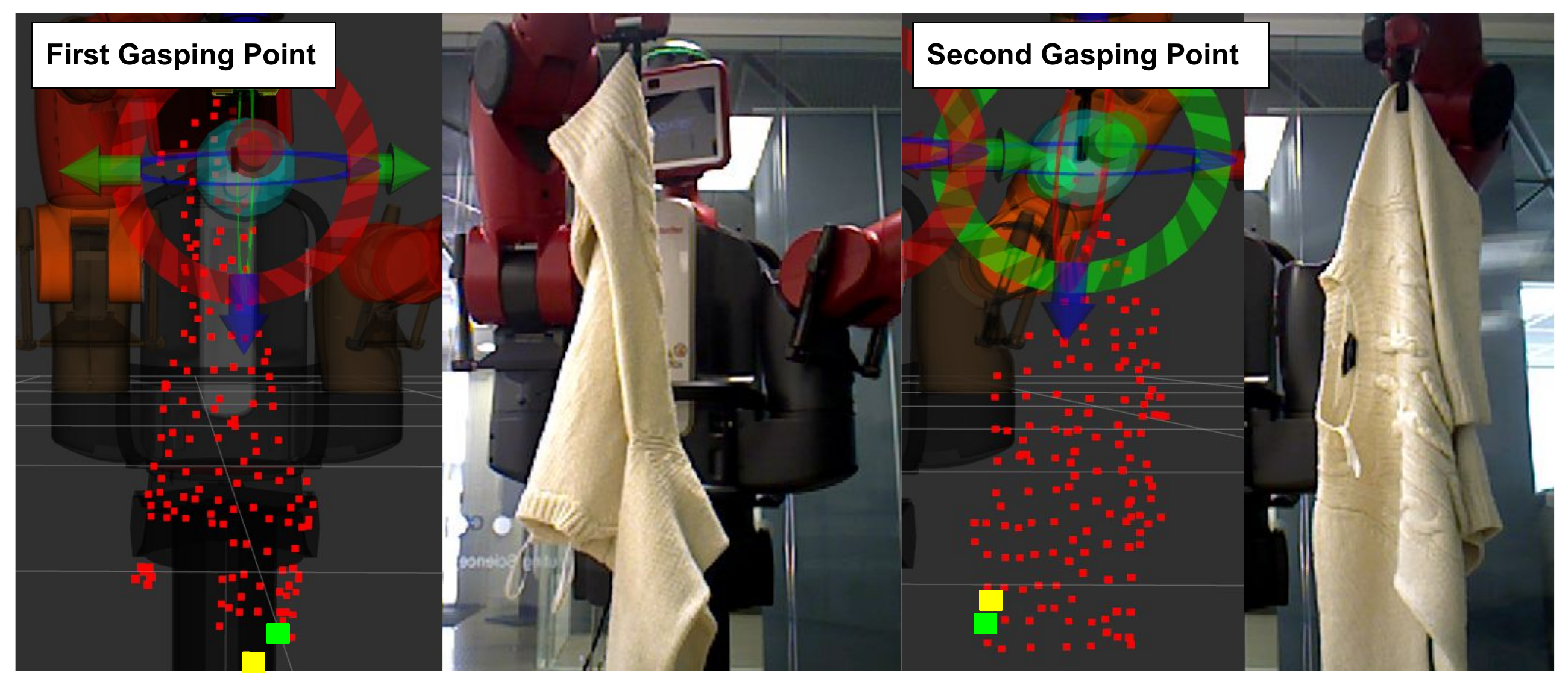}
    \caption{\textit{Grasping Points Location with Point Cloud}: the pre-designed grasping points need to be fine-tuned to ensure stable robotic grasps. We use a point cloud to fine-tune the pre-designed grasping points. (\textit{Top}: fine-tuning the pre-designed first grasping point of a sweater; \textit{Bottom}: fine-tuning the pre-designed second grasping point of the sweater; \textit{Yellow}: the pre-designed grasping points; \textit{Green}: the fine-tuned grasping points)}
    \label{fig:grasping_points_location_with_point_cloud}
\end{figure}

In our experiments, we do not need to identify local features or landmarks of garments as used in previous research (\cite{qian2020cloth, li2015regrasping, triantafyllou2022garment}); we only need to do visual servoing when \textit{known configurations} are recognised. Therefore, our data-centric approach does not require constructing models for garments and updating manipulation plans for unseen garments. Figure \ref{fig:grasping_points_location_with_point_cloud} shows an example of pre-designed grasping points (coloured in yellow) and fine-tuned grasping points (coloured in green) for the first and second grasping points (described in section \ref{subsec:pre-designed_manipulation_plans}) of a sweater. We can observe that the pre-designed grasping points are near to the garment but the fine-tuned grasping points can ensure stable grasps compared with pre-designed grasping points.

\subsection {Robotic Setup \label{subsec:robotic_setup}}
A Baxter dual-arm robot is employed for our experiments, controlled by MoveIt software. A four-legged table is placed in front of the Baxter robot to place garments in front of the robot. Two Xtion stereo cameras are set up facing the Baxter robot at two locations to enable wide- and narrow-field image capturing. Both cameras capture RGBD images (i) for predicting garment shapes and (ii) for recognising garment \textit{known configurations} and fine-tuning pre-designed points (Section  \ref{subsec:grasping_points_locating_with_point_cloud}). For predicting garment shapes, the garment similarity network (Section \ref{subsec:prior_knowledge_of_the_shapes_of_garments}) needs to observe how garments deform during grasping. Therefore, the camera must capture garment deformations as the robot it up; hence, this camera is placed at a larger distance from the Baxter robot. Conversely, for recognising the \textit{known configurations} of garments (Section \ref{subsec:robotic_garment_flattening_with_known_configuration_network}), the camera needs to capture the details of garments and is, thus, placed close to the Baxter robot. We implement our manipulation plans using the robot operating system (ROS, Kinect version) and the Ubuntu 16.04 environment, installed on a computer with a Core i7 CPU and an Nvidia 1080 Ti GPU. An \textit{openni2} \footnote{\url{http://wiki.ros.org/openni2_launch}} camera-controlling system controls the cameras. The sampling rate is 30 Hz for both cameras.

We use five shapes of garments with four different instances for each shape, totalling 20 garments tested. The shapes considered in this paper are jeans, shirts, sweaters, towels and t-shirts. Garments have different colours, textures and materials. Jeans are made of denim, shirts are made of polymers, and sweaters, t-shirts and towels are made of cotton. Experiments are conducted in a robot laboratory, where lighting conditions are set the same during each experiment. The laboratory is enclosed, which means that the external environment is stable and does not affect the experiments.

\section{Experiment Setup \label{sec:experiment_setup}}
\subsection {Database}
We used the database that we constructed in \cite{duan2022recognising}. For each grasping point (equivalent to each \textit{known configuration}), we captured 100 RGB and depth images. For each garment instance, the robot randomly grasps the garment and lifts it above the table ten times, totalling 19,269 images in our database. Note that 731 images had image artefacts or noise. Thus, we removed them from our database. Each image has a resolution of $256\times256$.

\subsection {KCNet training experiments}
As described in \ref{sec:methodology}, KCNet\footnote{Our implementation of KCNet can be found at \url{https://liduanatglasgow.github.io/Robotic_Garment_Flattening_with_Known_Configurations/}} consists of ResNet18 and three linear networks, where the features of predicted shapes and the images of \textit{known configurations} are concatenated and used for recognising \textit{known configurations}.  

We implement a leave-one-out cross-validation method in the experiments to validate our approach. Our KCNet is trained on a small database. Thus, implementing the leave-one-out cross-validation method ensures that our KCNet has a robust performance in recognising the \textit{known configurations} of garments. The database is divided into training and testing groups. The images from three of four garment instances for each shape serve as a training group, while the images of the remaining instance serve as the testing group. We ensure that the garments in testing groups never appear in training groups, which means testing garments are unseen by GarNet and KCNet. Therefore, four experiments have been conducted, where testing and training groups differ each time. We averaged the recognition accuracies across the four experiments, and the averaged value is used as the accuracy of the KCNet. We compare KCNet combined with GarNet in this paper with the KCNet in \cite{duan2022recognising}. We conduct an ablation study on recognition accuracy on depth, RGB, and RGBD images to investigate image formats' impacts on recognising \textit{known configurations}. We will use the KCNet combined with GarNet with the highest accuracy for our robotic garment flattening pipeline. We used an Adam optimiser to train KCNet, which is regulated by a learning-rate scheduler with a step size of eight epochs and a decay factor of $10^{-1}$. The initial learning rate for the optimiser is $10^{-3}$, the number of training epochs is 30 and the batch size is 64.

\subsection{Robotic Garment Flattening Pipeline \label{sebsec:robot_garment_flattening_pipeline_testing}}

We performed 20 garment flattening operations to validate our robotic garment flattening pipeline experimentally. For this, we recorded starting configuration states, final configuration states, times for each manipulation step, and success rates (the percentage of successful tests in all tests). A starting/final configuration state is computed as:
\begin{equation}
Starting \ State= \frac {S_{start}}{S_{goal}} \times 100 \%   
\end{equation}
\begin{equation}
Final \ State= \frac{S_{ending}}{S_{goal}} \times 100  \%  
\end{equation}

$S_{start}$ refers to the area in pixels of the starting configurations of garments, and $S_{ending}$, the area in pixels of the final configurations of garments. $S_{goal}$ represents the area in pixels of the goal configurations of garments. A goal configuration is a garment configuration when the garment is fully flattened (referenced in \cite{wu2019learning}). An example of goal configurations is shown in Figure \ref{fig:robotic_demonstration_examples}. To capture the starting, final and goal configuration, we positioned a camera at a fixed distance above the table, and captured images at each state and cropped them using the table's corners and edges as guidelines. 

\section{Experiment Results \label{sec:experiment_results}}

\subsection{KCNet training and ablation study results}

Tables \ref{tab:kcnet_training_results_without_prior_knowledge} and \ref{tab:kcnet_training_results_with_prior_knowledge} show the results for unseen garment instances of KCNets trained on different image types. We can observe that the KCNet trained on depth images ($92\%$ for the KCNet with prior knowledge of garment's shapes and $89\%$ for the KCNet without prior knowledge of garment's shapes) has the best performance compared to RGB ($78\%$ with prior knowledge of garment's shapes and $73\%$ without) and RGBD ($78.5\%$ with prior knowledge of garment's shapes and $81.5\%$ without). Depth images capture structural and spatial information about garments, compared to RGB images which capture texture information. Structural and spatial characteristics are similar between garments of similar shapes, while textures are easily affected by lighting conditions and the garment's colours that vary across different garments. Also, compared with KCNets without prior knowledge of garment's shapes \cite{duan2022recognising}, KCNets with prior knowledge of garment's shapes demonstrate better performance on all types of images, specifically on recognising \textit{known configurations} of shirts and sweaters. That is, shirts and sweaters share similar structures, thus, identifying differences between their \textit{known configurations} based only on their depth images is challenging. The latter is overcome by having prior knowledge of the garment's shape.

\begin{table*}[t]
    \centering
    \caption{KCNet ablation study: KCNet without prior knowledge of the garment's shape (Accuracy, unit: \%)}
    \label{tab:kcnet_training_results_without_prior_knowledge}
    \begin{tabular}{c|cccc|cccc|cccc}
    \hline
     SHAPE  & Exp 1 & Exp 2 & Exp 3 & Exp 4 & Exp 1 & Exp 2 & Exp 3 & Exp 4 & Exp 1 & Exp 2 & Exp 3 & Exp 4\\ \hline
     towel &94.4&96.4&93.1& 86.2 & 67.0& 55.9& 74.1& 64.5 & 71.8 & 79.7 & 85.2 & 76.7\\
     tshirt &86.8& 87.2 & 96.3& 94.7 & 70.8& 71.2 & 72.1 & 76.9 & 74.6  & 68.3 & 84.4 & 83.9\\
     shirt &78.2 & 80.3 &75.9 & 92.5 & 87.9 & 58.4 & 75.4& 91.4 & 87.9 & 58.4& 68.8 & 91.9 \\
     sweater &78.4& 85.4 &87.0 &86.2 & 54.6 & 42.0 & 68.1 & 85.2  & 52.7 & 51.6& 74.7 & 77.0 \\
     jean &99.3 & 95.8& 95.3 &99.1 & 76.4&87.8 & 87.0 & 98.8 & 87.5 & 92.0 & 97.2 & 99.5 \\
     \textit{average} & \textit{87.0} & \textit{89.0} & \textit{89.0} & \textit{92.0} &\textit{73.0} & \textit{62.0} & \textit{75.0}& \textit{83.0} & \textit{76.0} & \textit{70.0} & \textit{82.0}  & \textit{86.0} \\ \hline
     &\multicolumn{4}{c|}{Depth AVERAGE: \textit{89.0} } & \multicolumn{4}{|c|}{RGB AVERAGE: \textit{73.0}} & \multicolumn{4}{|c}{RGBD AVERAGE: \textit{78.5}}
     \\
     \hline
    \end{tabular}
\end{table*}

\begin{table*}[t]
    \centering
    \caption{KCNet ablation study: KCNet with prior knowledge of the garment's shape (Accuracy, unit: \%)}
    \label{tab:kcnet_training_results_with_prior_knowledge}
    \begin{tabular}{c|cccc|cccc|cccc}
    \hline
     SHAPE  & Exp 1 & Exp 2 & Exp 3 & Exp 4 & Exp 1 & Exp 2 & Exp 3 & Exp 4 & Exp 1 & Exp 2 & Exp 3 & Exp 4\\ \hline
     towel &93.8&96.5&92.4&87.5 &67.2&78.5&85.8&63.9 &74.2 &79.3 &84.0 & 65.6\\
     tshirt &93.9&93.1&94.4 & 95.0& 84.6& 71.7& 73.5& 79.8& 85.3 & 77.8 & 90.8 & 86.7\\
     shirt &90.1& 82.0&90.4& 96.2 & 90.5& 68.1& 72.8&95.2 &90.5 &73.2 &67.5 & 94.4 \\
     sweater &76.0&90.3&92.5&83.7 &54.0&56.3&81.2&71.0 &61.1 &67.2 & 77.0 & 84.3 \\
     jean &98.5&94.2&94.5&98.9 & 82.0&89.9& 86.5&98.0&90.4 &87.5 & 95.0 & 98.5\\
     \textit{average} & 91.0&91.0&93.0& 92.0 &77.0& 72.0& 80& 82.0 & 81.0 & 77.0 & 82.0 & 86.0\\ \hline
     &\multicolumn{4}{c|}{Depth AVERAGE: 92.0} & \multicolumn{4}{|c|}{RGB AVERAGE: 78.0} & \multicolumn{4}{|c}{RGBD AVERAGE: 81.5}
     \\
     \hline
    \end{tabular}
\end{table*}

\subsection{Robotic Garment Flattening Results}

Table \ref{tab:robotic_garment_flattening_testing_results} shows the starting configuration states, running times for each step, final configuration states and success rates. On average, the Baxter robot requires 29 seconds to predict the garment shapes with GarNet. For garment flattening, Baxter needs, on average, 79.8 seconds to grasp the first grasping point, 60.2 seconds to grasp the second grasping point, and 52.6 seconds to stretch and flatten a garment. Compared to the state-of-art \cite{yan2020learning}, where the robot needs approximately 15 minutes to fold a piece of garments, our approach has a faster manipulation operation. Traditional approaches update robotic manipulation plans in real-time, which requires costly computations. Also, updating robotic manipulation plans may result in robots' unnecessary actions, taking additional time.

On average, the starting configuration states are $27.80\%$ (ref. Section \ref{sebsec:robot_garment_flattening_pipeline_testing}), while it increases to $80.1\%$ for final configuration states. The results show that the robot successfully flattened garments and reached a high average final configuration state (80.1\%) with our approach.  We found that jeans, shirts, sweaters and t-shirts have better final configuration states than towels. When Baxter finds the second grasping point for flattening the towels, it sometimes grasps the incorrect corners because the towels are symmetrical, resulting in final configuration states lower than other garments.

Table \ref{tab:comparison_with_state_of_art} shows a comparison between our approach with state-of-art. Sun \textit{et al.} \cite{sun2015accurate} was  one of the first to demonstrate dual-arm flattening of towels and achieved a dual-arm success rate of $46.3\%$; their approach required approx. 20 minutes to flatten a squared towel with one or two folds. Li \textit{et al.} \cite{li2015regrasping} and Martin \textit{et al.} \cite{5509439} achieved high manipulation success rates; however, they required re-grasping garments to ensure that the robot recognised the pose (equivalent to the \textit{known configuration} in this paper). Their approaches, therefore, required longer manipulation time. Li \textit{et al.} \cite{li2015regrasping} achieved a lower pose estimation accuracy than our approach because they trained their network with simulated data. Martin \textit {et al.} \cite{5509439}, Yan \textit{et al.} \cite{yan2020learning} and Sun \textit{et al.} \cite{sun2015accurate} only tested towels in their experiments, while we used five different shapes of garments. Overall, our experiments achieved the fastest manipulation and second highest \textit{known configuration} recognition accuracy but have a lower success rate than other approaches.

The average success rate for our flattening approach is $66.7\%$. We can observe that jeans and towels have the lowest success rates ($57.1\%$). Jeans are heavier than other garments, so they sometimes slip from the robotic grippers and result in failures. As we mentioned, towels are sometimes grasped from wrong grasping points because they are symmetric, resulting in failure cases. In future research, we plan to use grippers suited for manipulating garments (e.g. \cite{sun2015accurate}) for grasping and flattening jeans and better manipulation plans for grasping and flattening towels. Figure \ref{fig:robotic_demonstration_examples} shows three successful garment flattening operations of three shapes (a shirt, a jean and a sweater). A video demonstration can be found at \url{https://youtu.be/j7yEbJcAgDM}. 

\begin{table*}[t]
    \centering
    \caption{Robotic Garment Flattening Results}
    \begin{tabular}{c|c|ccccc|c|c}
    \hline
    SHAPE & \textbf{Starting State} & \textit{Shape Prediction} & \textit{Grasping Point 1} & \textit{Grasping Point 2} & \textit{Flattening} &  \textit{Total Time} & \textbf{Final State} & Success Rate \\\hline
    Jeans  &   \textbf{40.51\%} & \textit{26.0s} & \textit{55.5s}  & \textit{50.0s} & \textit{72.5s} & \textit{204.0s} & \textbf{87.75\%}  & 57.1\% \\ 
    Shirt &   \textbf{27.04\%} & \textit{30.0s} & \textit{71.0s}  & \textit{47.5s} & \textit{42.0s} & \textit{190.5s} & \textbf{96.30\%}  & 66.7\% \\ 
    Sweater & \textbf{25.98\%} & \textit{28.8s} & \textit{86.0s}  & \textit{62.0s} & \textit{66.0s} & \textit{242.8s} & \textbf{76.65\%}  & 80.0\%) \\ 
    Towel &   \textbf{23.85\%} & \textit{30.0s} & \textit{82.5s}  & \textit{71.0s} & \textit{41.0s} & \textit{224.5s} & \textbf{65.44\%}  & 57.1\% \\ 
    T-shirt &  \textbf{21.61\%} & \textit{30.0s} & \textit{104.0s} & \textit{70.5s} & \textit{41.5s} & \textit{246.0s} & \textbf{74.06\%}  & 80.0\% \\ \hline
    Average & \underline{\textbf{27.80\%}} & \underline{\textit{29.0s}} & \underline{\textit{79.8s}} & \underline{\textit{60.2s}} & \underline{\textit{52.6s}} & \underline{\textit{221.6s}} & \underline{\textbf{80.10\%}} & \underline{66.7\%} \\ \hline
    \end{tabular}
    \label{tab:robotic_garment_flattening_testing_results}
\end{table*}

\begin{table}[tbhp]
    \centering
    \caption{Comparison with state-of-art (N/A means that the value is not reported in the paper.)}
\begin{tabular}{|c|c|c|c|}
    \hline
    Method  &  Estimation Accuracy & Running Time & Success Rate\\ \hline
    Sun \cite{sun2015accurate} & N/A & 1,200 & 46.3\% \\ \hline
    Li \cite{li2015regrasping} & 83\% & N/A & 80.0\% \\ \hline
    Yan \cite{yan2020learning} & N/A & 900s & N/A \\ \hline
    Maitin \cite{5509439} & \textbf{96\%} & 1478 s & \textbf{81\%} \\ \hline
    Ours & 92\% & \textbf{221.6s} & 66.7\% \\ \hline
\end{tabular}
    \label{tab:comparison_with_state_of_art}
\end{table}


\section{Conclusion \label{sec:conclusion}}
In this paper, we have proposed an effective robotic garment flattening pipeline where a Baxter robot predicts garment shapes, recognises their hanging state (\textit{known configurations} of garments) based on their shapes and depth images, and uses pre-designed manipulation plans to flatten them. We found that our approach achieved an accuracy of $92\%$ for recognising \textit{known configurations}, and the robot takes, on average, $29$ seconds to predict the garment shapes and $192.6$ seconds on average to manipulate and flatten garments.

In this paper, we only use garments from the five shapes because manipulation plans are designed based on these five shapes. Garments with an unknown shape (for example, shorts) can not be manipulated, limiting our approach to specific shapes of garments, and this is an open problem in deformable object manipulation. We plan to implement a reinforcement learning approach \cite{pore2020simple} to improve the success rate of garment flattening, where the robot learns basic skills which can be composed to manipulate any garment shape. Another limitation of our flattening pipeline is that we use two cameras to recognise the garment's shapes and known configurations. In future work, we plan to use only one camera for both shape and \textit{known configuration} predictions and investigate active vision approaches to control the camera's viewpoint.


\bibliographystyle{IEEEtran}
\bibliography{references.bib}

\end{document}